# SURVEY ON VARIOUS GESTURE RECOGNITION TECHNIQUES FOR INTERFACING MACHINES BASED ON AMBIENT INTELLIGENCE

Harshith.C[1], Karthik.R.Shastry[2], Manoj Ravindran[3], M.V.V.N.S Srikanth[4], Naveen Lakshmikhanth[5]

Department of Information Technology
Amrita Vishwa Vidyapeetham
Coimbatore

{[1]harshith.c, [2]karthikrshaz ,[3]manojr.ravindran ,[4]mvvnssrikanth ,[5]naveen.lakshmikhanth}@gmail.com

## ABSTRACT

*Gesture recognition is mainly apprehensive on analyzing the functionality of human wits. The main goal of gesture recognition is to create a system which can recognize specific human gestures and use them to convey information or for device control. Hand gestures provide a separate complementary modality to speech for expressing ones ideas. Information associated with hand gestures in a conversation is degree, discourse structure, spatial and temporal structure. The approaches present can be mainly divided into Data-Glove Based and Vision Based approaches. An important face feature point is the nose tip. Since nose is the highest protruding point from the face. Besides that, it is not affected by facial expressions. Another important function of the nose is that it is able to indicate the head pose. Knowledge of the nose location will enable us to align an unknown 3D face with those in a face database.*

*Eye detection is divided into eye position detection and eye contour detection. Existing works in eye detection can be classified into two major categories: traditional image-based passive approaches and the active IR based approaches. The former uses intensity and shape of eyes for detection and the latter works on the assumption that eyes have a reflection under near IR illumination and produce bright/dark pupil effect. The traditional methods can be broadly classified into three categories: template based methods, appearance based methods and feature based methods. The purpose of this paper is to compare various human Gesture recognition systems for interfacing machines directly to human wits without any corporeal media in an ambient environment.*

## KEYWORDS

Ambient Intelligence Human wits, Hand gesture, Nose tips, Eye detection.

## 1. INTRODUCTION

Hand gestures provide a separate complementary modality to speech for expressing ones ideas. Information associated with hand gestures in a conversation is degree, discourse structure, spatial and temporal structure. The approaches present can be mainly divided into Data-Glove Based and Vision Based approaches. The Data-Glove based methods use sensor devices for digitizing hand and finger motions into multi-parametric data. The extra sensors make it easy to collect hand configuration and movement. However, the devices are quite expensive and bring much cumbersome experience to the users. In contrast, the Vision Based methods require only a camera, thus realizing a natural interaction between humans and computers without the use of any extra devices. These systems tend to complement biological vision by describing artificial vision systems that are implemented in software and/or hardware. This poses a challenging problem as these systems need to be background invariant, lighting insensitive, person and camera independent to achieve real time performance.

An important face feature point is the nose tip. This is because the nose is the highest protruding point from the face. Besides that, it is not affected by facial expressions. Another important function of the nose is that it is able to indicate the head pose. Knowledge of the nose location will enable us to align an unknown 3D face with those in a face database. Some 2D works include training the computer to detect the nose using Support Vector Machine or by using contrast values and edge detection to locate the nose changes. Using 3D images, one of the methods used was to take horizontal slices of the face and then draw triangles on the slices. The point with the maximum altitude triangle will be considered the nose tip.

Eye detection is divided into eye position detection and eye contour detection. Existing works in eye detection can be classified into two major categories: traditional image-based passive approaches and the active IR based approaches. The former uses intensity and shape of eyes for detection and the latter works on the assumption that eyes have a reflection under near IR illumination and produce bright/dark pupil effect. The traditional methods can be broadly classified into three categories: template based methods, appearance based methods and feature based methods.

## 2. RELATED WORK

Eyes and Nose are the most salient and robust features on human faces. The precise detection of eyes and the nose tip had been a crucial step in many face-related applications. A handful of hardware solutions came its way through which included, entering data into the computer by pressing some sort of switches, infrared emitters and reflectors attached to the user's glasses, head band, or cap, transmitter over the monitor. All these solutions were not properly adaptable as some people felt uncomfortable in wearing helmets, glasses, mouth sticks or other interfacing devices. In the later years, development in technology resulted in locating the eyes and nose computationally. The various approaches include the following:

In the template based methods, a generic eye model, based on the eye shape, was designed at first [12]. Template matching is then used to search the image for the eyes. While these methods can detect eyes accurately, they are normally time-consuming. In order to improve the accuracy, these methods have to match the whole face with an eye template pixel by pixel. If we are not aware of the size of eyes then, the matching process with eye templates of different sizes is repeated. Solutions to improve the efficiency of template matching method focus on two points:

  i. Reducing the area in the face image for template matching
  ii. Cutting down the times of this type of matching

In addition, one can evaluate the size of eye template according to the size of these two regions. In other words, profiting from possibility of evaluating the size of eyes, this algorithm performs the template matching just once [12]. A template based approach in which the convex shape of the nose was used to identify the nose region. It involved calculating the local search area which was used as a template. The template was scanned through the image and the pixel 'x' which had maximum correlation was identified. The position of best match was refined using Evident Based Convolution Filter.

Appearance based methods: Appearance based methods detect eyes based on their photometric appearance. These methods usually need to collect a large amount of training data, representing eyes of different individuals under different illumination conditions.

This data is used to train some classifier, and detection is achieved by classification [15] [17].

Texture based methods: The candidates of eyes are estimated by eye filters based on texture information. Human eyes usually include important information of a face, compared to the other facial features. The appearance of eyes variously changes.

Pixel-pattern-based texture feature (PPBTF) - It is constructed from one pattern map. A gray scale image is first transformed into a pattern map in which edge and background pixels are classified by pattern matching with a given set of M pattern templates B that reflect the spatial features of images. For each pixel (x, y) in a gray scale image I, let $z_i$ be the inner product of its S × S neighbour block b within the pattern templates.

Feature based methods: In which the characteristics (such as edge and intensity of iris, the colour distributions of the sclera and the flesh) of the eyes were explored to identify some distinctive features around the eyes. Although these methods were efficient, they lacked accuracy for the images which did not have high contrast. For example, these techniques mistook eyebrows for eyes.

A few systems which detect facial features computationally include the following:

Colour Data Model: The human facial features such nose tip and the nose ridge were identified using a shading technique on the colour data of the image. But, the highly sensitive nature of the nose region to light made the system disadvantageous.

Effective Energy: An Effective Energy (EE) term was defined for nose tip, based on the assumption that the nose tip is the highest local point and has peaked cup like shape. This method had the sufficient accuracy result in case the complete shape of nose tip can be obtained. This is a feature based approach.

User Interface development requires a sound understanding of human hand's anatomical structure in order to determine what kind of postures and gestures are comfortable to make. Although hand postures and gestures are often considered identical, the distinctions between them need to be cleared. Hand posture is a static hand pose without involvement of movements. For example, making a fist and holding it in a certain position is a hand posture. Whereas, a hand gesture is defined as a dynamic movement referring to a sequence of hand postures connected by continuous motions over a short time span, such as waving good-bye. With this composite property of hand gestures, the problem of gesture recognition can be decoupled into two levels- the low level hand posture detection and the high level hand gesture recognition. In vision based hand gesture recognition system, the movement of the hand is recorded by video camera(s). This input video is decomposed into a set of features taking individual frames into account. Some form of filtering may also be performed on the frames to remove the unnecessary data, and highlight necessary components. One of the earliest model based approaches to the problem of bare hand tracking was proposed by Rehg and Kanade [19]. In [20] a model-based method for capturing articulated hand motion is presented. The constraints on the joint configurations are learned from natural hand motions, using a data glove as input device. A sequential Monte Carlo tracking algorithm, based on importance sampling, produces good results, but is view-dependent, and does not handle global hand motion.

The organization of the rest of this paper is as follows. Section III highlights the various aspects of hand posture and gesture recognition technology. Section IV discusses the available algorithms for nose detection and tracking, discussing their advantages and shortcomings. Section V describes various application areas of eye recognition and tracking and points out the open issues, and Section VI concludes the paper.

## 3. HAND DETECTION AND RECOGNITION

### 3.1 Hidden Markov Models

This method (Hidden Markov Model [1]) deals with the dynamic aspects of gestures. Gestures are extracted from a sequence of video images by tracking the skin-colour blobs corresponding to the hand into a body– face space centered on the face of the user. The goal is to recognize two classes of gestures: deictic and symbolic. The image is filtered using a fast look–up indexing table

33

of skin colour pixels in YUV colour space. After filtering, skin colour pixels are gathered into blobs. Blobs are statistical objects based on the location (x,y) and the colourimetry (Y,U,V) of the skin colour pixels in order to determine homogeneous areas. A skin colour pixel belongs to the blob which has the same location and colourimetry component. Deictic gestures are pointing movements towards the left (right) of the body–face space and symbolic gestures are intended to execute commands (grasp, click, rotate) on the left (right) of shoulder.

### 3.2 YUV Colour Space and CAMSHIFT Algorithm

This method deals with recognition of hand gestures. It is done in the following five steps.

1. First, a digital camera records a video stream of hand gestures.

2. All the frames are taken into consideration and then using YUV colour space skin colour based segmentation is performed. The YUV colour system is employed for separating chrominance and intensity. The symbol Y indicates intensity while UV specifies chrominance components.

3. Now the hand is separated using CAMSHIFT [2] algorithm .Since the hand is the largest connected region, we can segment the hand from the body.

4. After this is done, the position of the hand centroid is calculated in each frame. This is done by first calculating the zeroth and first moments and then using this information the centroid is calculated.

5. Now the different centroid points are joint to form a trajectory .This trajectory shows the path of the hand movement and thus the hand tracking procedure is determined.

### 3.3 Using Time Flight Camera

This approach uses x and y-projections of the image and optional depth features for gesture classification. The system uses a 3-D time-of-flight (TOF) [3] [4] sensor which has the big advantage of simplifying hand segmentation. The gestures used in the system show a good separation potential along the two image axes. Hence, the projections of the hand onto the x- and y-axis are used as features for the classification. The region of the arm is discarded since it contains no useful information for the classification and due to strong variation between human beings. Additionally, depth features are included to distinguish certain gestures: gestures which have same projections, but different alignments.

The algorithm can be divided into five steps:

1. Segmentation of the hand and arm via distance values: The hand and arm are segmented by an iterative seed fill algorithm.

2. Determination of the bounding box: The segmented region is projected onto the x- and y-axis to determine the bounding box of the object.

3. Extraction of the hand.

4. Projection of the hand region onto the x- and y-axis.

### 3.4 Naïve Bayes' Classifier

This method is an effective and fast method for static hand gesture recognition. This method is based on classifying the different gestures according to geometric-based invariants which are obtained from image data after segmentation; thus, unlike many other recognition methods, this method is not dependent on skin colour. Gestures are extracted from each frame of the video, with a static background. The segmentation is done by dynamic extraction of background pixels according to the histogram of each image. Gestures are classified using a weighted K-Nearest Neighbors Algorithm which is combined with a Naïve Bayes [5] approach to estimate the probability of each gesture type. When this method was tested in the domain of the JAST Human Robot dialog system, it classified more than 93% of the gestures correctly.

This algorithm proceeds in three main steps. The first step is to segment and label the objects of interest and to extract geometric invariants from them. Next, the gestures are classified using a K-nearest neighbor algorithm with distance weighting algorithm (KNNDW) to provide suitable data for a locally weighted Naïve Bayes' classifier. The input vector for this classifier consists of invariants of each region of interest, while the output is the type of the gesture. After the gesture has been classified, the final step is to locate the specific properties of the gesture that are needed for processing in the system—for example, the fingertip for a pointing gesture or the center of the hand for a holding-out gesture.

### 3.5 Vision Based Hand Gesture Recognition

In vision based hand gesture recognition system [6], the movement of the hand is recorded by video camera(s). This input video is decomposed into a set of features taking individual frames into account. The hands are isolated from other body parts as well as other background objects. The isolated hands are recognized for different postures. Since, gestures are nothing but a sequence of hand postures connected by continuous motions, a recognizer can be trained against a possible grammar. With this, hand gestures can be specified as building up out of a group of hand postures in various ways of composition, just as phrases are built up by words. The recognized gestures can be used to drive a variety of applications.

The approaches to Vision based hand posture and gesture recognition

(i)  3D hand model based approach

(ii) Appearance based approach

### 3.5.1 3D Hand Model Based Approach

Three dimensional hand model based approaches rely on the 3D kinematic hand model with considerable DOF's, and try to estimate the hand parameters by comparison between the input images and the possible 2D appearance projected by the 3D hand model. Such an approach is ideal for realistic interactions in virtual environments. This approach has several disadvantages that have kept it from real-world use. First, at each frame the initial parameters have to be close to the solution, otherwise the approach is liable to find a suboptimal solution (i.e. local minima). Secondly, the fitting process is also sensitive to noise (e.g. lens aberrations, sensor noise) in the imaging process. Finally, the approach cannot handle the inevitable self-occlusion of the hand.

### 3.5.2 Appearance Based Approach

This method use image features to model the visual appearance of the hand and compare these parameters with the extracted image features from the video input. Generally speaking, appearance based approaches have the advantage of real time performance due to the easier 2D image features that are employed. There have been a number of research efforts on appearance based methods in recent years. A straightforward and simple approach that is often utilized is to look for skin coloured regions in the image. Although very popular, this has some drawbacks like skin colour detection is very sensitive to lighting conditions. While practicable and efficient methods exist for skin colour detection under controlled (and known) illumination, the problem of learning a flexible skin model and adapting it over time is challenging. This only works if we assume that no other skin like objects is present in the scene. Another approach is to use the

International Journal of Computer Science & Engineering Survey (IJCSES) Vol.1, No.2, November 2010

eigenspace for providing an efficient representation of a large set of high-dimensional points using a small set of basis vectors.

## 4. NOSE DETECTION AND RECOGNITION

### 4.1 Using Geometric Features and FFT

In this method, for each pixel in the image the generalized eccentricities ε0 and ε2 are calculated and are mapped to the feature space. In feature space, noses are characterized by bounding box which is learned from a set of labeled training data. During classification, a given pixel is said to belong to a tip of a nose if and only if it is mapped into this bounding box in feature space. For FFT based approach the background was set to a constant values (the max value appearing in the foreground); for NFFT [7] approach the background pixels was simply discarded. To segment the foreground object, an adaptive threshold to amplitude range is applied.

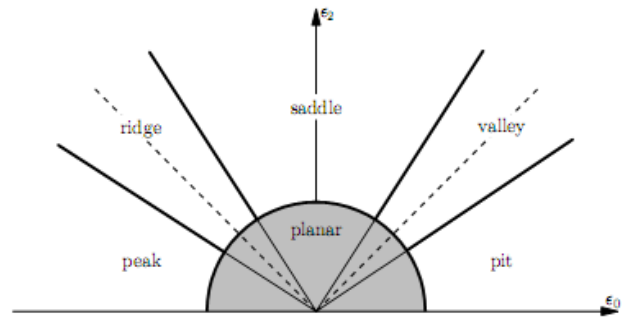

Figure 1 : Feature Space

The Figure 1 represents the discrimination of the six surface types- pit, peak, saddle, valley, ridge, and planar within the feature space spanned by ε0 and ε2.

### 4.2 Using Hough Transform

In this method (Hough Transform [8]), the face vertexes in proximity of the found sphere are extracted, and magnifying the extracted cloud of a factor of 10, in order to center the sphere on the nose. The mean value of normal vectors of each face vertex is evaluated with the purpose of obtaining information about face orientation in the three-dimensional space. The center of the sphere is then moved by the direction evaluated, and proportionally to the sphere beam value. In this way we collapse the entire sphere in a point that represents the nose-tip. The point that is obtained may not be present among the vertex of the examined face. So the last step consists to research the closest face vertex to the nose-tip previously extracted. This vertex represents the real nose-tip.

### 4.3 Effective Energy Calculation

Since the Nose is the protruding area on the facial front, this method uses Edge Detection techniques [9] to identify potential nose tip candidates by investigating the neighbouring pixels. The nose tip is then identified from the potential candidates by their effective energy, mean and variance values obtained from Principal Component Analysis Technique. This method has the ability to locate the nose tip in Frontal as well as Non-Frontal Faces.

### 4.4 Structural and Holistic Features

To locate the nose tip template of the nose is used. Then, the template matching technique based on correlation [10] is employed and the point, at which the correlation between the template and the sub-image is the maximum, is defined as nose tip. But, the template matching is a time consuming process and to reduce the search time we first locate the eye-line. The ideal eye-line is the horizontal line which passes through the pupil of the eyes. To locate the eye-line, the upper half of the image is considered and then the horizontal projection of the Sobel gradient is computed. Then, the t-horizontal lines having highest gradients are selected. From, these t- lines the middle one represent the approximate eye-line. The template matching process starts from the eye-line and continues up to, say, row r, where r is half of the height of the original image.

### 4.5 Colour Data Model
The human facial features such nose tip and the nose ridge were identified using a shading technique on the colour data of the image. But, the highly sensitive nature of the nose region to light made the system disadvantageous.

### 4.6. Template Based Method
In this approach the convex shape of the nose is used to identify the nose region. It involves calculating the local search area which was used as a template. The template was scanned through the image and the pixel and one which had maximum correlation was identified. The position of best match was refined using Evident Based Convolution Filter [11].

## 5. EYE DETECTION AND TRACKING

### 5.1 Template Based Method
In the template based methods [12], a generic eye model, based on the eye shape, is designed at first. Template matching is then used to search the image for the eyes. While these methods can detect eyes accurately, they are normally time-consuming. In order to improve the accuracy, these methods have to match the whole face with an eye template pixel by pixel. If we are not aware of the size of eyes then, we need to repeat the matching process with eye templates of different sizes. That is to say, the template matching has to be performed several times. In fact, this method at first detects the two rough regions of eyes in the face using a feature based method. Thus the following template matching will be performed only in these two regions which are much smaller than the whole face. Solution to improve the efficiency of template matching method focuses on two points - reducing the area in the face image for template matching and cutting down the times of this type of matching. In addition, the size of eye template can be evaluated according to the size of these two regions. In other words, profiting from possibility of evaluating the size of eyes, this algorithm performs the template matching just once. Altogether, the method combines the accuracy of template based methods and the efficiency of feature based methods.

### 5.2 Appearance Based Method
Appearance based methods [13] detect eyes based on their photometric appearance. These methods usually need to collect a large amount of training data, representing eyes of different individuals under different illumination conditions. This data is used to train some classifier, and detection is achieved by classification.

### 5.3 Pixel-Pattern-Based Texture Feature (PPBTF)
PPBTF [14] is firstly proposed in. It is constructed from one pattern map. A gray scale image is first transformed into a pattern map in which edge and background pixels are classified by pattern matching with a given set of M pattern templates B that reflect the spatial features of images. For each pixel $(x, y)$ in a gray scale image I, let $z_i$ be the inner product of its $S \times S$ neighbour block b with the pattern templates. Then the pixel $(x, y)$ in the pattern map P is assigned a number k such that $z_k = \max(z_1, z_2, \ldots, z_m)$. Therefore, the pixel values in a pattern map represent the pattern classes of pixels in the original gray scale image.

The feature model comes as follows: Suppose the number of patterns is M, then the pixel value P(x, y) in the pattern map P is in a range of [1, M]. For each pixel (i, j), the features in a window S1×S1 can be generated by,

$$f_1(i,j) = \sum_{x=i-(S1-1)/2}^{i+(S1-1)/2} \sum_{y=j-(S1-1)/2}^{j+(S1-1)/2} h_i(x,y)$$

Equation 1

Where the function h is a binary function defined by,

$$h_i(x,y) = \begin{cases} 1, & P(x,y) = 1 \\ 0, & \text{otherwise} \end{cases}$$

Thus, the feature gives the number of the pixels belonging to the pattern, and the feature vector is constructed.

## 5.4 Cascade Classifier

Cascade of classifier [15] achieves increased detection performance while radically reducing computation time. The key insight is that smaller, and therefore more efficient, boosted classifiers can be constructed which reject many of the negative sub-windows while detecting almost all positive instances. Simpler classifiers are used to reject the majority of sub-windows before more complex classifiers are called upon to achieve low false positive rates. The stage in cascade is constructed by Ad boost. (Figure 2).

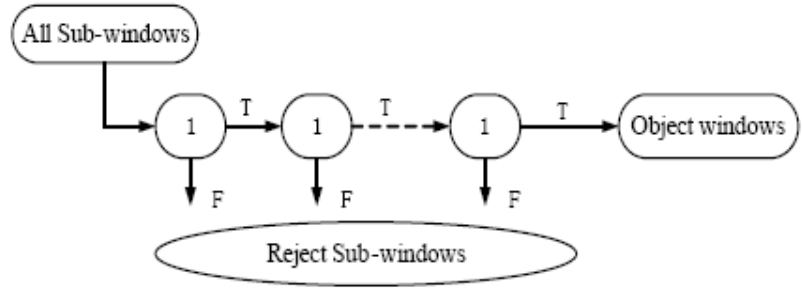

Figure 2: The structure of cascade classifier

After the detection of cascade classifier, some errors could arise, such as eyebrows, mouth, larger or smaller eyes, to exclude these patches, some other method to classify the eye and non-eye patches that come out from the cascade classifier is necessary, and this is done using PPBTF.

## 5.5 Principal Component Analysis

Pattern templates represent the spatial features in an image and reflect that how the value of 1 pixel depend on those of its neighbours. The process of obtaining the templates is as follows: First for a given image, an S × S image block is denoted as a neighbor vector, and N numbers of S × S image blocks are randomly created. If the image numbers are N, and an (S × S) × (M × N) matrix A would come out, then PCA [16] will be computed using the matrix A. After PCA analysis eigenvectors are obtained, and some of the eigenvectors are selected as the template. The first one corresponding to the largest eigenvalue is a Gaussian low-pass filter, and the others are derivative filters. Except the first basis function, the others can be used like gradient filters for pattern matching. In the transformation of a gray scale image (24×12) eye to a pattern map, each pixel is assigned the index of the templates which best matches its 3×3 neighbour block. Since the templates resemble derivative operators, the value of a pixel in the pattern map represents the edge pattern class of its neighbour block.

### 5.6 Ada boost and SVM Classifier

First Ada boost [17] is used to select features instead of being a classifier. A subset of the features is selected by Ada boost at each step; the chosen feature is uncorrelated with the output of the previous features. Because the selected features are less than the total features, the speed is improved. SVM classifiers are trained on the features selected by AdaBoost. A 5-fold cross-validate scheme is used in the operation. The best SVM [17] parameters are gained by a method called grid search. RBF kernel function is used for the SVM. The main parameters of SVM are penalty C and Sigma (parameter of RBF kernel). To improve the speed, both of them are chosen between 25~220 in the experiment. The facial image is put into the cascade classifier using rectangle features, and the result patches of the classifier are scaled and transformed into a pattern map using PCA basis functions to form the features which will be classified by Ad boost and SVM classifier.

### 5.7 Colour Based Detection

Using colour characteristics is a useful way to detect eyes. Two maps are made according to its components and merge them to obtain a final map. Candidates are generated on this final map. An extra phase is applied on candidates to determine suitable eye pair. The extra phase consists of flexible thresholding and geometrical tests. Flexible thresholding makes generating candidates more carefully and geometrical tests allow proper candidates to be selected as eyes, first build two separate eye maps [18] from facial image, EyeMapC from the chrominance components and EyeMapL from the luminance component. These two maps are then combined into a single eye map. The main idea of EyeMapC is based on characteristics of eyes in YCbCr colour space which demonstrates that eye regions have high Cb and low Cr This formula is designed to brighten pixels with high Cb and low Cr values.

$$\text{EyeMapC} = \frac{1}{3}\left\{\left((C_b)^2 + (\overline{C_r})^2 + \left(\frac{C_b}{C_r}\right)\right)\right\}$$

Equation 2

Since the eyes usually contain both dark and bright pixels in the luma component, grayscale morphological operators (e.g., dilation and erosion) can be designed to emphasize brighter and darker pixels in the luma component around eye regions. This is basically the construction of the EyemapL.

$$\text{EyeMapL} = \frac{Y(x, y) \oplus g(x, y)}{Y(x, y) \otimes g(x, y)}$$

Equation 3

After constructing EyeMapC and EyeMapL, we multiply them to obtain the final 'EyeMap', i.e., EyeMap= (EyeMapC) AND (EyeMapL)

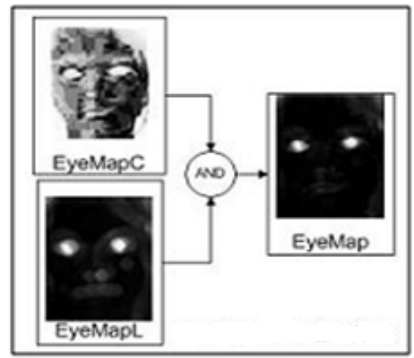

Figure 3: Eye Map Construction

## 6. CONCLUSION

Based on the observations regarding Hand Detection and Tracking, we can conclude that using YUV Skin Color Segmentation followed by CAMSHIFT algorithm will help in the effective detection and tracking as the centroid values can easily be obtained by calculating the moments at each point, later we could combine Hidden Markov Training for further applications. It is better when compared to Time-Flight Camera where one has to find the bounding box and then use Iterative Seed Fill algorithm.

For Nose Detection and Tracking system, the results obtained using Edge Detection Techniques were comparatively better, and it also detects the nose for faces at various angles. The success rate using this method for frontal faces is about 93% and for non frontal faces it is about 68%. This shows that this particular method is suitable to be implemented in an automatic 3D face recognition system.

For the analysis of Eye Detection and Tracking, it would be efficient to implement a method which would be a combination of texture based and colour based eye detection methods as colour plays a major role in both texture recognition of facial regions and later finding the accurate location of eyes using eye maps. Individually the efficiencies of texture based and color based detection methods are 98.2% and 98.5% respectively, but after incorporating the proposed method the efficiency could be enhanced to greater extent.

### ACKNOWLEDGEMENTS

We would like to thank our parents for the blessings that they showered on us and the continuous support we received. We would also like to thank our project mentor Mr.T.Gireesh Kumar and our Department Head Mr.K.Gangadharan for their invaluable assistance throughout the work.

International Journal of Computer Science & Engineering Survey (IJCSES) Vol.1, No.2, November 2010

**AUTHORS PROFILE**

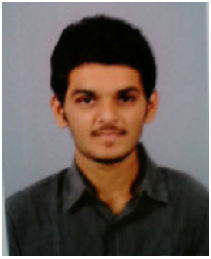

**Harshith C** is currently pursuing his B.Tech in Information Technology from Amrita Vishwa Vidyapeetam, Coimbatore. His major research interests are Human Computer Interaction (Artificial Intelligence) and Computer Programming.

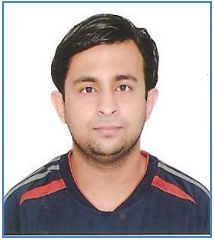

**Karthik R Shastry** is currently pursuing his B.Tech in Information Technology from Amrita Vishwa Vidyapeetam, Coimbatore. His major research interests are Artificial Intelligence and Application and Software Development.

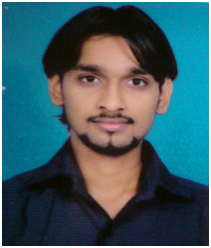

**Manoj Ravindran** is currently pursuing his B.Tech in Information Technology from Amrita Vishwa Vidyapeetam, Coimbatore. His major research interests are Artificial Intelligence and Software Programming.

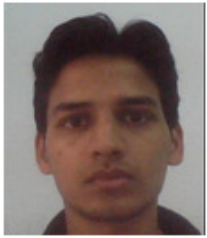

**MVVNS Srikanth** is currently pursuing his B.Tech in Information Technology from Amrita Vishwa Vidyapeetam, Coimbatore. His major research interests are Artificial Intelligence and Computer Networks.

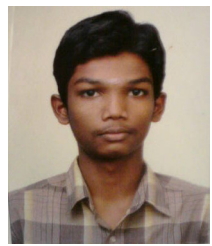

**Naveen Lakshmikhanth** is currently pursuing his B.Tech in Information Technology from Amrita Vishwa Vidyapeetam, Coimbatore. His major research interests are Artificial Intelligence and Networking.